\documentclass[10pt,twocolumn,letterpaper]{article}

\usepackage{cvpr}              

\usepackage{graphicx}
\usepackage{bbm}
\usepackage{amsmath}
\usepackage{amssymb}
\usepackage{booktabs}

\usepackage{paralist}

\usepackage{soul}
\usepackage{wrapfig}

\usepackage[accsupp]{axessibility}  

\usepackage[pagebackref,breaklinks,colorlinks]{hyperref}

\usepackage[capitalize]{cleveref}
\crefname{section}{Sec.}{Secs.}
\Crefname{section}{Section}{Sections}
\Crefname{table}{Table}{Tables}
\crefname{table}{Tab.}{Tabs.}

\newcommand{\mypartitle}[2][2.3]{\vspace*{-#1 ex}~\\{\noindent {\bf #2}}}

\usepackage{xspace}
\newcommand{\mbf}{\mathbf}

\newcommand{\pixellocation}{\ensuremath{\mbf{p}}\xspace}
\newcommand{\eyelocation}{\ensuremath{\mbf{p}_{eye}}\xspace}
\newcommand{\predictedgazelocation}{\ensuremath{\mbf{p}_{gaze}}\xspace}
\newcommand{\predictedgazeTwoD}{\ensuremath{\mbf{g}_{2D}}\xspace}
\newcommand{\gtgazeTwoD}{\ensuremath{\mbf{g}^{gt}_{2D}}\xspace}

\newcommand{\heatmap}{\ensuremath{\mbf{H}}\xspace}
\newcommand{\inoutindicator}{\ensuremath{o}\xspace}

\newcommand{\headcrop}{\ensuremath{\mbf{I}_{head}}\xspace}
\newcommand{\headmask}{\ensuremath{\mbf{I}_{mask}}\xspace}

\newcommand{\inputimage}{\ensuremath{\mbf{I}}\xspace}
\newcommand{\inputraw}{\ensuremath{\mbf{I_{raw}}}\xspace}

\newcommand{\modality}{\ensuremath{m}\xspace}
\newcommand{\imagemodality}{\ensuremath{\mbf{I_{\modality}}}\xspace}
\newcommand{\featuremap}{\ensuremath{\mbf{F}}\xspace}
\newcommand{\transformedfeaturemap}{\ensuremath{\mbf{T}}\xspace}
\newcommand{\embedding}{\ensuremath{e}\xspace}
\newcommand{\weight}{\ensuremath{w}\xspace}

\newcommand{\gazeconeimage}{\ensuremath{\mbf{I}_{co}}\xspace}
\newcommand{\coneaperture}{\ensuremath{\alpha_{co}}\xspace}

\newcommand{\gazesubnetwork}{\ensuremath{\mathcal{G}}\xspace}
\newcommand{\featuresubnetwork}{\ensuremath{\mathcal{F}}\xspace}
\newcommand{\attentionnetwork}{\ensuremath{\mathcal{A}}\xspace}
\newcommand{\regressionnetwork}{\ensuremath{\mathcal{R}}\xspace}
\newcommand{\inoutnetwork}{\ensuremath{\mathcal{O}}\xspace}

\def\cvprPaperID{10} 
\def\confName{CVPR}
\def\confYear{2022}

\begin{document}

\title{A Modular Multimodal Architecture for Gaze Target Prediction:\\ Application to Privacy-Sensitive Settings \\[-4mm]}

\author{Anshul Gupta, Samy Tafasca, Jean-Marc Odobez\\
Idiap Research Institute, Martigny, Switzerland\\
Ecole Polytechnique Fédérale de Lausanne, Switzerland\\
{\tt\small \{agupta, stafasca, odobez\}@idiap.ch}
}
\maketitle

\begin{abstract}
\vspace {-3mm}
   Predicting where a person is looking is a complex task, requiring to understand not only the person's gaze and scene content, but also the 3D scene structure and the person's situation (are they manipulating? interacting or observing others? attentive?) to detect obstructions in the line of sight or apply attention priors that humans typically have when observing others.
   In this paper, we hypothesize that identifying and leveraging such priors can be better achieved through the exploitation of explicitly derived multimodal cues such as depth and pose. We thus propose a modular multimodal architecture allowing to combine these cues using an attention mechanism.
   The architecture can naturally be exploited in privacy-sensitive situations such as surveillance and health, where personally identifiable information cannot be released.
   We perform extensive experiments on the GazeFollow and VideoAttentionTarget public datasets, obtaining state-of-the-art performance and demonstrating very competitive results in the privacy setting case. 
   \footnote{Code is available at \url{https://github.com/idiap/multimodal_gaze_target_prediction}}
\end{abstract}

\vspace {-4mm}


\section{Introduction}
\label{sec:intro}

\begin{figure}[th]
  \includegraphics[width=\columnwidth]{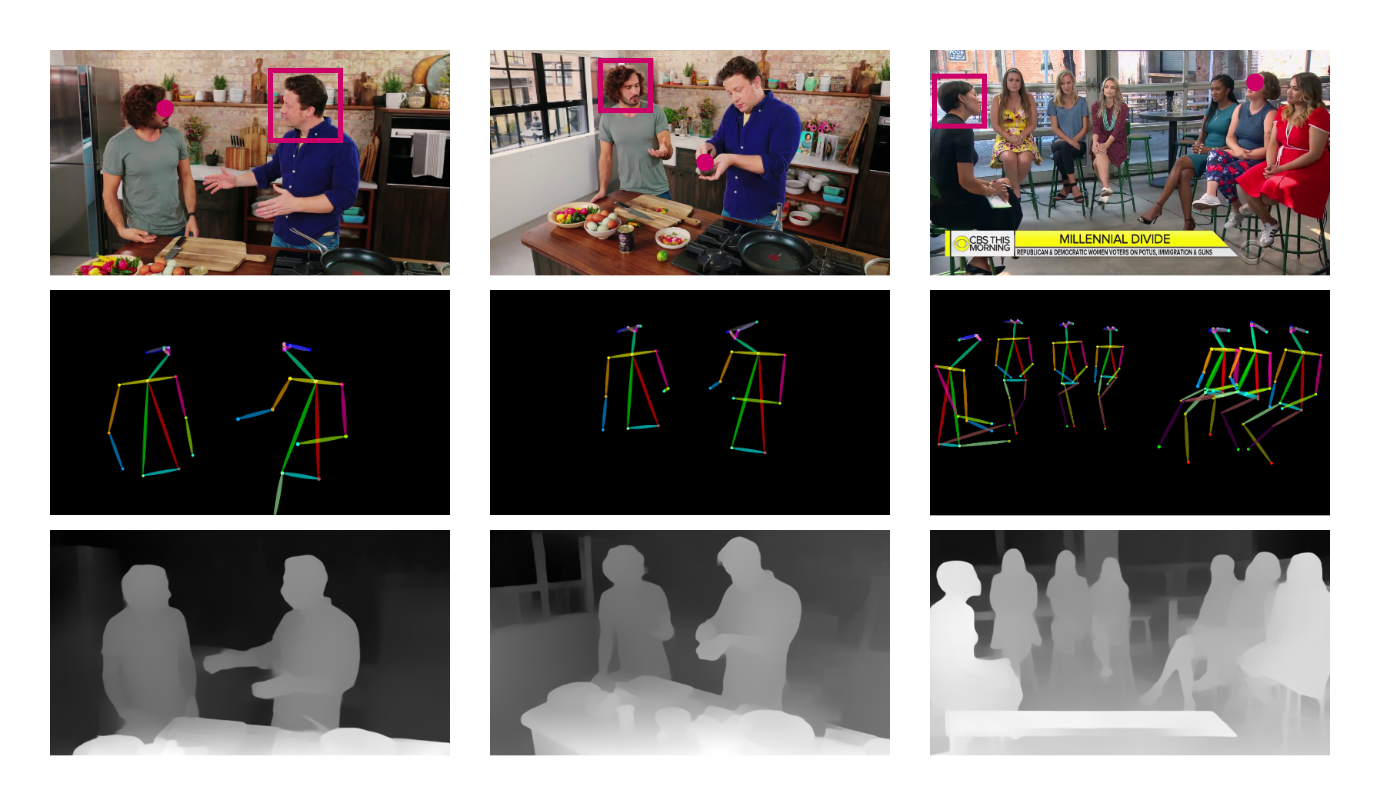}
  \vspace*{-8mm}
  \caption{Sample images where depth and pose information can be useful to infer the gaze target.
    Left: the pose can indicate that people are interacting, while depth
    helps to rule out salient objects in the background.
    Middle: manipulation activities where knowledge of the hands can be useful.
    Right: depth allows to filter out the potential face candidates in the back.
    }
  \vspace*{-3mm}
\label{fig: sample-images-modalities}
\end{figure}


As an indicator of attention, gaze is an important cue which can reveal considerable information about a person's behavior or state of mind.
In this regard, identifying the gaze of people in visual data finds applications in many domains,
like in the retail industry to understand consumer behaviour \cite{tomas2021goo},
in sociology for assessing  social gaze behaviours such as shared attention \cite{fan2018inferring_videocoatt}
or higher constructs like dominance within meetings \cite{Ba:MLMI:2006,Hung:ICMI:2008},
or in human-robot interaction for communication analysis \cite{sheikhi2015_vfoa}.


In recent years, there has been an increased body of work devoted to gaze analytics.
One research direction focuses on improving the raw gaze prediction, defined as the 3D angular values representing the 3D line of sight.
These methods typically use the eyes~\cite{Yu_2018_gaze} or the face of a person~\cite{kellnhofer2019gaze360} as input.
Another research line addresses the identification of the visual focus of attention (VFOA), \ie the visual target a person
is looking at \cite{sheikhi2015_vfoa,Otsuka2018,Bai2019}. 
Such a task is challenging, as it requires not only to capture the body, head and potentially the eyes of the person of interest
to infer their attention, but also the understanding and monitoring of the scene containing the gaze targets.
Due to this constraint, traditional methods usually perform the task in fixed environments
relying on multi-camera set-ups \cite{Otsuka2018,Bai2019} or prior knowledge of scene location \cite{sheikhi2015_vfoa}.
This creates challenges for applying the methods and models to unseen environments.

With a focus on generalization, Recasens et al.~\cite{recasens2017following} proposed to address
the VFOA task using a single image, by formulating it as the prediction of the
image 2D location of the gaze target looked at by a person of interest in the image.
They proposed a model that implicitly learns what the salient elements in a scene
are, and how to combine them with the attention evidence obtained from the gaze information inferred
from the person's head crop image.
Relying on a large annotated dataset, good results were obtained, with the advantage that such an approach can, in theory,
be applied to any arbitrary scene.
This work has also been extended in many ways \cite{lian2018believe, chong2018connecting,zhao2020learning,guan2020_pose,nan2021predicting,jin2021multi}.
%
In this paper, we address this 2D target prediction task and investigate different factors which can contribute
to the success and understanding of the inference process,  as motivated below.


\mypartitle{Motivations.}
Several cues modulate gaze following behavior in humans, such as saliency and social context \cite{shepherd2010following}. Transposing this idea to images, we argue that predicting the gaze target of a person in an image can benefit from leveraging such information.
This includes inferring the general gaze direction of the person,
and identifying the salient items (potential VFOA targets) located in their field of view, like objects and faces.
To further remove ambiguities, humans usually rely on priors about gaze behaviours, which depend
on understanding  the 3D structure of the scene (to check visibility factors)
as well as the current context (task performed, ongoing interactions, intentions, past actions, $\ldots$).
While the information to reach this level of understanding and apply the right prior
for inferring the gaze is directly available in the image,
one may wonder whether explicitly providing visual cues and modalities would ease and improve the inference of gaze target.

In this regard, we study the use of explicit depth and pose information for improving attention inference,
as illustrated in Fig. \ref{fig: sample-images-modalities}.
Indeed, depth information gives the model an idea about object shapes and the 3D scene structure. 
It allows for understanding whether a person is looking to the foreground or background,
and can help resolve ambiguities along the line of sight when the depth does not match.
On the other hand, pose provides accurate information
about the locations of body parts related to attention such as hands and faces
which are common gaze targets during interactions and manipulation activities.
Pose also provides information about the physical state and activity of the person(s),
which can help decide which categories of potential gaze targets the network needs to focus on.  

Using images with visible faces can be an issue in privacy-sensitive scenarios,
either when obtaining training data, or at inference time. 
Surveillance and health are typical  applications. 
For instance, it has been shown that reduced eye contact and shared attention are early warning signs for autism in children\cite{Zwaigenbaum_autism}.
However, due to their sensitive nature, raw videos (even with faces blurred since we are interested in gaze)
are not available for public access, making it difficult to develop and test models.
On the other hand, pose and depth data do not contain identifying information, so they can be shared. Hence, we develop models which use only pose and depth data, and evaluate their performance in this paper. 
%

We also study the benefit of other technical elements.
The first one is resolution.
Indeed, gaze target localization is  a dense prediction task similar to pose landmarks estimation:
our output is a heatmap corresponding to the probability of a point being the gaze target.
Current approaches~\cite{lian2018believe, chong2020dvisualtargetattention, Fang_2021_CVPR_DAM}
typically use ResNet style architectures where the spatial resolution of features is greatly reduced before
being upsampled again for the final prediction.
Instead, we adopt a Feature Pyramid Network approach~\cite{lin2017fpn}
which includes skip connections during the upsampling process
to preserve spatial information and demonstrate improved results.
Secondly, in current methods, gaze information is often merged with the input image content to
infer the gaze of a person. This early fusion requires the full image to be (re)processed for each person.
We investigate whether a late fusion approach can be adopted (fusing gaze information with feature maps)
and show that it does not achieve the same level of performance.

\mypartitle{Approach and contributions.}
In summary, we address the gaze target location prediction task and make the following contributions:
\begin{compactitem}
\item we propose a modular multimodal gaze prediction architecture with end-to-end training and an attention scheme to combine the saliency features of image, pose and depth modalities;
\item we investigate the use of only pose and depth information in our setting to allow its usage in privacy-sensitive settings; 
\item we propose the use of a Feature Pyramid Network regression  scheme and
  show that preserving spatial information is important due to the nature of the task. 
\end{compactitem}
We conduct experiments on the  GazeFollow~\cite{recasens2017following}
and VideoAttentionTarget~\cite{chong2020dvisualtargetattention} public benchmarks,
and show that we obtain state-of-the-art results using all modalities,
and that competitive results can be obtained using only depth and pose information which is of
interest for privacy-sensitive applications.


\section{Related Work}
\label{sec:related-work}

This paper relates mainly to the problems of gaze target prediction and, to a minor extent, data anonymization.
Below is a review of works in these topics. 

\subsection{Gaze Target Prediction}

When accurate gaze trackers were not available, VFOA was often inferred from head pose using behavioral models~\cite{Stiefelhagen1999}.
Inference mechanisms like GMM, HMM, or Dynamical Bayesian Networks~\cite{Otsuka2006, Ba2008} were used to estimate the VFOA directly from the head pose,
and potentially using  other contextual information \cite{Gorga2010} which can act as priors
on the VFOA like the speaking status, the speech semantic content~\cite{sheikhi2015_vfoa,Otsuka2018},
or modeling interactions and the joint VFOA of all participants~\cite{Ba2011, Masse2018}.
%
Nevertheless, with the recent improvement of gaze estimation, even simple frame-based geometrical models were shown to be effective to estimate VFOA~\cite{Yucel2013}.
Recent models used deep networks such as CNNs and RNNs, resulting in further improved performance~\cite{siegfried2021_vfoa}.
However all these methods typically rely on some prior knowledge about the scene  structure and hence can not generalize to arbitrary settings.

To address generalization,
Recasens et al.~\cite{nips15_recasens} formulated the problem as
the inference of the 2D image position corresponding to the location of the scene target a person (in the image) is looking at. 
They proposed a CNN model combining the information from two branches,
a saliency branch which processes the scene and a gaze branch analysing
the head crop of the person of interest.
Most models that followed relied on a similar two branch architecture.
For instance, Chong et al.~\cite{chong2018connecting} extended the model to also predict whether a person is looking inside or outside the frame.
Lian et al.~\cite{lian2018believe} predicted a 2D gaze vector from the gaze branch and used it to generate explicit gaze cones which
were then concatenated along with the input image for inference. We follow a similar idea to generate a gaze cone which is concatenated
with each modality, and we investigate the privacy-sensitive setting. 
Drawing inspiration from works in human pose estimation, 
Zhao \etal~\cite{zhao2020learning} proposed an interesting method which learned to predict the line of sight as well as infer the attention 'landmark', 
and demonstrated improved results over the other baselines.
Other works proposed to process multiple people together~\cite{jin2021multi} or use temporal information~\cite{chong2020dvisualtargetattention}
using an LSTM module at the bottleneck, but in the latter case, results were not improved much compared to the frame-based case.

There has also been some work exploring the use of multimodal information.
Guan et al.~\cite{guan2020_pose} used the pose of the person of interest to supplement the gaze branch in cases where the face is not visible.
In the approach of Nan et al.~\cite{nan2021predicting}, authors aim to merge (task driven) top-down attention
with bottom-up features (flow and pose) to derive the gaze target. 
While they perform a similar late fusion of features across modalities,
our fusion mechanism operates at a much higher resolution and relies on an attention mechanism.
In addition, their overall method (with top-down features) is quite different and was applied in a specific setting. 
Fang et al.~\cite{Fang_2021_CVPR_DAM} used depth to potentially disambiguate
attention targets by inferring whether a person is looking toward their foreground or background, 
obtaining the best results reported so far on the GazeFollow and VideoAttentionTarget datasets. Recently, Hu et al.~\cite{hu2022we} used depth information to perform gaze target prediction in 3D.
As far as we are aware, ours is the first work to use both pose and depth information, and to study
the privacy preserving situation.

Finally, there are other works addressing tasks related to gaze following. 
This includes predicting gaze target objects \cite{tomas2021goo},
detecting whether two people are looking at each other \cite{Marin-Jimenez_2019_CVPR} or
recognizing shared attention behavior \cite{fan2018inferring_videocoatt}.
However, due to their aims, these works differ substantially from the study we conduct here.

\subsection{Data Anonymization}

The typical approaches for anonymizing face data include techniques such as blurring and pixelation~\cite{Boyle2000_blur}\cite{Gross2009_blur}. However, these methods may not remove privacy-sensitive information~\cite{Newton2005_notblur,Neustaedter2006_notblur} and may instead remove critical information for the downstream task. 
More recent methods use generative adversarial methods~\cite{hukkelaas2019deepprivacy} to alter the face. However, these methods are not suited for our task as changes to facial features can affect 
the gaze information.
Instead, we propose the use of pose information (facial landmarks) 
to predict the gaze direction.
This removes identifying information while still giving 
us a good approximation of a person's gaze direction. 
This was recently demonstrated by Belkada \etal~\cite{belkada2021pedestrians}, 
who showed that the head and body pose alone could be used to predict the "eye contact" of people with a camera sensor placed on a car. 
Our results further confirm this hypothesis in more general scenes.

\section{Model Architecture}

\begin{figure*}
\centering
\includegraphics[width=0.9\textwidth]{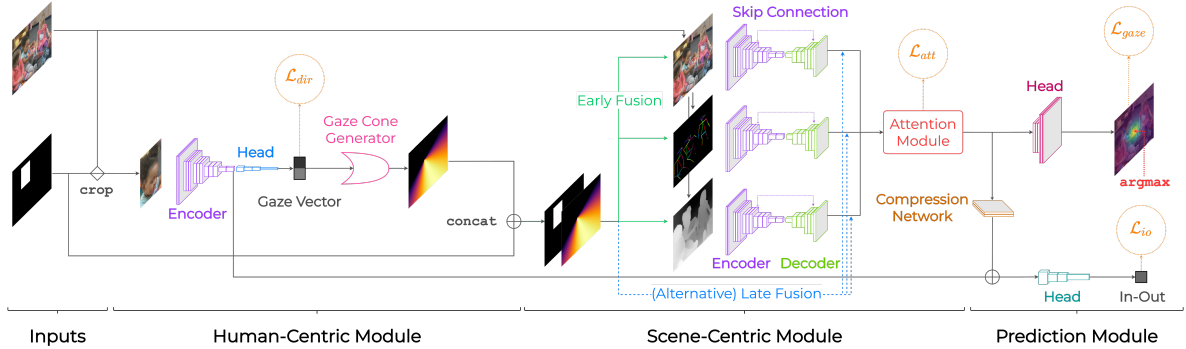}
\vspace*{-2mm}
\caption{Overview of our proposed architecture.
  Given an input image and a target subject's head location, we first extract depth and pose maps from the image using off-the-shelf pre-trained models.
  Next, the Human-Centric module takes the person's head crop as input and predicts a 2D gaze vector which is used to generate a gaze cone image.
  Then, the Scene-Centric module processes the original image, the depth image and the pose map in order to produce modality saliency feature maps
  (using modality-specific encoder-decoder feature extractors) which are fused by an Attention module.
  The resulting saliency map is used by the Prediction module to regress a gaze heatmap, and optionally predict an in-vs-out gaze classification score.
}
\vspace*{-2mm}
\label{fig:full-model}
\end{figure*}

\subsection{Approach overview}
\label{sec:approach-overview}

An overview of our system is illustrated in Figure~\ref{fig:full-model}.
It takes as input an image or a video frame, a set of derived modality images, and the head bounding box of a target person. The output is a gaze heatmap \heatmap where the location of the maximum value corresponds to the desired gaze prediction \predictedgazelocation.

Our network architecture consists of 3 modules.
The first one is a \emph{Human-Centric module} whose goal is, given the head crop of a person,
to predict a gaze cone
representing their visual field of view,
\ie the set of pixel locations where the person might be looking.
The second one is a \emph{Scene-Centric module} which is fed the image, the person's location (head mask) and the
gaze cone in order to generate a feature saliency map \featuremap highlighting possible gaze target locations.
The last one is a \emph{Prediction module} comprising two heads: one for inferring the gaze heatmap,
and the second one for  predicting 
whether the gaze target is located within the frame.
These components are detailed below.

\subsection{Human-Centric Module}
\label{sec: human-centric-module}

In this module, a sub-network \gazesubnetwork takes the head crop image $\headcrop$ of the target person as input 
and predicts a normalized 2D gaze vector $\predictedgazeTwoD = \gazesubnetwork(\headcrop)$.
This gaze vector is used by a \emph{gaze cone generator} to produce a gaze cone image $\gazeconeimage$.
Finally, the gaze cone image is concatenated with the binary head mask of the target person $\headmask$ and passed
to the Scene-Centric branch for further processing.

\mypartitle{Gaze Cone Generator.}
The gaze cone is a way to modulate the image information appearing in the gaze direction of the person. 
It is encoded as an image in order to be consistent with the rest of the architecture.
The gaze cone generator produces   \gazeconeimage  by drawing a cone from the subject's eyes location $\eyelocation$ 
(i.e. eye mid-point if available from the pose modality; otherwise, using a prototypal location in the head bounding box)
along the direction of $\predictedgazeTwoD$.
To account for uncertainties in gaze prediction, the cone has an aperture of $\coneaperture$ (set to $\pi$ in practice), and the intensity decays the farther we are from the gaze direction angle-wise. 
Specifically, the value at each pixel location $\pixellocation$  is scored according to the cosine similarity between
the predicted gaze vector and the eye-to-target direction.
(see example in Fig.~\ref{fig:full-model}).
Formally:
%
\begin{equation}
\begin{aligned}
    \forall \pixellocation =(i,j) \text{ where } (i,j) \in [1, \ w] \times [1, \ h], \\
    \gazeconeimage(\pixellocation) = \max\left(0, \text{cos}(\predictedgazeTwoD, \pixellocation  - \eyelocation) \right)
\end{aligned}
\end{equation}
Note that given this definition, the gaze cone generator is  differentiable, allowing to train our architecture end-to-end.

\subsection{Scene-Centric Module}
\label{sec:scene-centric-module}

In the Scene-Centric module, the input image $\inputimage$ is first transformed using different networks (see implementation details) into a set of modality images $\imagemodality$, where $\modality \in \{raw, pose, depth \}$ and $\inputraw = \inputimage$ by definition. These modalities are passed through feature extractors to produce feature maps, which are then fused using an attention mechanism to create a single combined feature map.

\mypartitle{Feature Extractors.}
A set of modality-specific feature extractors $\featuresubnetwork_{\modality}$ are used to compute feature maps $\featuremap_{\modality}$
to encode the person-specific salient regions of the scene according to the input modality.
Thus, each feature extractor $\featuresubnetwork_{\modality}$ processes its corresponding modality
$\imagemodality$ concatenated with the output of the Human-Centric module, so that we have:
\begin{equation}
  \label{eq:featuremaps}
  \featuremap_{\modality} = \featuresubnetwork_{\modality}(\imagemodality, \gazeconeimage, \headmask)
\end{equation}
%
%
Note that the concatenation can be seen as an early fusion scheme, whereas an alternative (so far less successful) approach
consists in fusing the Human-Centric module information later at the feature level (see late fusion experiments). 
While multiplication is a more straightforward way to fuse the modality image and the gaze cone,
in practice it produced worse results,  probably because it performs a hard decision based on potentially
inaccurate gaze  direction predictions.
This is particularly the case when the subject’s head is facing backwards and the gaze vector is more difficult to estimate.
Concatenation on the other hand, allows the model to make that decision later in the processing.


Regarding the network, we used a typical image-to-image approach, relying on an encoder-decoder architecture.
However, in contrast to previous works which simply upsample the lowest resolution representation
produced by the encoder \cite{lian2018believe, chong2020dvisualtargetattention},
we used  skip connections from different intermediate representations (at different resolutions) to their corresponding decoder representations in the style of a Feature Pyramid Network \cite{lin2017fpn}.
This architectural choice aims to retain information from higher resolution representations,
which is important in dense prediction tasks, and further evidenced by our experiments.
%

\mypartitle{Attention Module.}
Its goal  is to perform a soft-selection of the most appropriate input modality given the scene.
It takes as input the set  of feature maps $\featuremap_{\modality} \in \mathbb{R}^{w \times h \times d_{\modality}}$
and produces a single combined feature map $\featuremap$, which we use to predict the outputs.
Concretely, it performs fours steps:
\begin{compactenum}
\item Each feature map $\featuremap_{\modality}$ is passed through a modality-specific convolution layer 
  to produce a transformed feature map $\transformedfeaturemap_{\modality}$.
  %
  %
\item Each map $\transformedfeaturemap_{\modality}$ is passed through a network $\attentionnetwork_{\modality}$ consisting of three strided convolution layers
  followed by a global max pooling to generate an embedding vector $\embedding_{\modality}$.
  All embeddings are then concatenated to form the global embedding $\embedding$.
\item The global embedding is passed through a projection layer $P$ followed by a softmax operation to get the attention weights:
  $\{\weight_{\modality}\} = \text{softmax}(P(e))$.
\item Finally, the output is computed as the weighted sum of the transformed feature maps:
  $\featuremap = \sum_{\modality} \weight_{\modality} \transformedfeaturemap_{\modality}$.
\end{compactenum}
This loosely resembles the self-attention mechanism in a transformer \cite{vaswani2017attention}:
the transformed feature maps $\transformedfeaturemap_{\modality}$ act as the values,
whereas the attention weights $\weight_{\modality}$ simulate a dot product between an implicit query and a set of keys.

In addition, this attention mechanism allows us to use a variable number of modalities during inference because
the model can simply assign a weight of $0$ when a modality is absent.
To encourage this behaviour, we perform modality dropout during training,
\ie we randomly provide a white noise image instead of the dropped modality,
and use an attention loss for supervision (see Section~\ref{sec:loss}).


\subsection{Prediction  Module}
\label{sec: attention-module}

This module uses the feature map to predict the quantity of interest: 
a gaze heatmap $\heatmap$,
and a binary In-Out flag $\inoutindicator$ indicating whether the gaze target is inside or outside the image.
It comprises two parts, which are explained below.

\mypartitle{Gaze prediction head.}
The gaze target heatmap $\heatmap$ is regressed from the combined feature map $\featuremap$
using a prediction decoder  $\regressionnetwork$ that consists of an analytic upsampling followed by a set of convolution layers:
\begin{equation}
    \heatmap = \regressionnetwork (\featuremap )
\end{equation}
The location where the heatmap is maximal is then used as the gaze target prediction.

\mypartitle{In-Out prediction head.}
In general, we want to predict whether the person is looking at a scene location which is visible in the image or not.
This is important as we do not want to use the gaze  target prediction when a person is looking outside the frame.
To accomplish this, we attach an In-Out network prediction head $\inoutnetwork$ which takes as input the feature map $\featuremap$
resulting from the attention step as well as a gaze embedding  $\embedding_{gaze}$ coming from the human centric module (see Fig.~\ref{fig:full-model}):
\begin{equation}
    \inoutindicator  = \inoutnetwork(\featuremap,  \embedding_{gaze} )
\end{equation}
More precisely:
first, the map $\featuremap$ is passed through a network having the same architecture as $\attentionnetwork_{\modality}$ to produce a scene embedding
$\embedding_{scene}$ which is concatenated with the gaze embedding $\embedding_{gaze}$ and fed into an In-Out predictor consisting of 2 linear layers followed by
a sigmoid activation.

\subsection{Loss}
\label{sec:loss}

The complete model is trained end-to-end using a combination of four losses:
\begin{compactenum}
\item {\bf Gaze loss $\mathcal{L}_{gaze}$}. It measures the error in gaze location prediction, which is done
by computing the pixel-wise L2 loss between the predicted  heatmap $\heatmap^{pred}$
and the ground truth gaze target heatmap $\heatmap^{gt}$, 
%
defined as a gaussian blob centered on the ground-truth location.
\item {\bf Gaze direction loss $\mathcal{L}_{dir}$}.
  The goal of this loss is to better constrain the learning of the Human-Centric module.
  This is achieved by maximizing the cosine of the angle between the predicted 2D gaze vector \predictedgazeTwoD
  and the ground truth vector \gtgazeTwoD, which we derive from the ground-truth gaze point.
\item {\bf In-Out loss  $\mathcal{L}_{io}$}.
  We use a standard binary cross entropy loss to measure whether a person is looking inside or outside the image frame.
\item {\bf Attention loss (modality drop) $\mathcal{L}_{att}$}.
  This loss aims to supervise the Attention module (Section \ref{sec:scene-centric-module}).
  The idea is to push the attention weight $\weight_{\modality}$ of a dropped modality \modality towards 0.
  Formally, the loss is  defined as:
  $\mathcal{L}_{att} = \sum_{\modality} \weight_{\modality} . \mathbbm{1}_{\modality \in \text{dropped}}$, where $\mathbbm{1}$
  is an indicator variable and 'dropped' is the list of dropped  modalities.
\end{compactenum}

\vspace{3mm}

The final loss is a linear combination of the four losses:
\vspace{-4mm}
\begin{equation}
    \mathcal{L} = \lambda_{gaze} \mathcal{L}_{gaze} + \lambda_{dir} \mathcal{L}_{dir} + \lambda_{io} \mathcal{L}_{io} + \lambda_{att} \mathcal{L}_{att}
\end{equation}

\subsection{Implementation Details}

\mypartitle{Modality extraction.}
The pose maps are extracted using HRFormer \cite{yuan2021hrformer}, a mix between HRNet \cite{wang2020deep} and the standard transformer architecture. On the other hand, the depth maps are extracted using MiDaS \cite{ranftl2019midas}, a strong monocular depth estimator. Pose maps are represented as RGB images of skeletons of the people in the frame, where different colors denote the different limbs and keypoints.

\mypartitle{Feature extraction networks $\featuresubnetwork_{\modality}$.}
The feature extractors in the scene branch use backbones chosen from the EfficientNet family \cite{tan2019efficientnet} because of their ability to scale capacity and expressive power without incurring a significant cost in terms of size.
Specifically, the image encoder is an EfficientNet-B1 (7.8M parameters) while the depth and pose encoders use an EfficientNet-B0 (5.3M parameters).
Input modalities are resized to $224 \times 224$ and fed to the EfficientNet backbones which compute different intermediate feature representations at resolutions
between $56 \times 56$ and $7 \times 7$.
These are used in the residual connections of the Feature Pyramid Network decoder
to produce the feature maps $\featuremap_{\modality}$ at resolution $56 \times 56$.

\mypartitle{Gaze subnetwork \gazesubnetwork}.
The Human-Centric branch, on the other hand, uses a ResNet-18 backbone (11M parameters) \cite{he2016deep_resnet}
equipped with a custom 2D gaze prediction head. This sub-network operates on the head crop of the target subject, resized to $224 \times 224$.

\mypartitle{Prediction module.}
Throughout the prediction module (cf. Figure \ref{fig:full-model}),
the feature maps $\featuremap_{\modality}$, $\transformedfeaturemap_{\modality}$, and \featuremap
are maintained at a resolution of $56 \times 56$, except in
the regression sub-network,  where $\featuremap$
is first upsampled to $64 \times 64$ (which is also the size of the predicted gaze heatmap) before going through different convolution layers.
The embedding vectors $\embedding_{gaze}$ (i.e. from the head crop encoder), $\embedding_{\modality}$ (within the attention
module) and $\embedding_{scene}$  (within the In-Out prediction head)  each have a size of $512$.

\section{Experiments}

\subsection{Experimental protocol}

We use two datasets for our experiments, and rely on standard metrics and protocols for evaluation.

\label{sec: gazefollow}
\label{sec: videoatt}

\mypartitle{Datasets.}
The first dataset is the {\em GazeFollow} dataset~\cite{nips15_recasens}.
It comprises a curated set of images from popular image datasets. It was initially annotated with the 2D gaze target location, eye location, and head bounding box for most people in the images.
Later, Chong et al.~\cite{chong2018connecting} extended these labels with indications of whether the gaze target of a person is located inside or outside the image.
Overall, the dataset contains annotations for around 130k people in 122k images. 
The test set consists of 4782 people (all looking inside the image) whose gaze was annotated by 10 annotators. 

The second one is the {\em VideoAttentionTarget} dataset~\cite{chong2020dvisualtargetattention}. 
It contains 1331 video clips collected from 50 shows on YouTube. 
The annotation comprises head bounding boxes and either the 2D gaze target location or whether the attention target is outside the frame.
The training and test sets contain respectively around 131k  and 33k bounding boxes. 
In general, the VideoAttentionTarget dataset 
has higher resolution frames and more close up and front-facing 
views of people compared to GazeFollow.

\mypartitle{Training Protocol.} 
%
%
To generate the ground-truth gaze heatmaps $\heatmap^{gt}$, 
we place a Gaussian centered on the ground-truth gaze point 
and use a $\sigma=3$ standard dev. (at the $64\times64$ heatmap resolution). 
In terms of training, the backbone of the sub-network in the Human-Centric branch is pre-trained on the Gaze360 dataset \cite{kellnhofer2019gaze360} to predict a 3D gaze vector, while the backbones of the feature extractors in the Scene-Centric branch are pre-trained on ImageNet \cite{russakovsky2015imagenet}. 
To train the multimodal models, we first train the individual modalities separately (see Sec.~\ref{sec:testedmodels}), and initialize their multimodal counterparts with the learned weights. Further, following the training protocol of \cite{Fang_2021_CVPR_DAM}, all experiments on VideoAttentionTarget use models initialized with weights learned from GazeFollow. For VideoAttentionTarget, we subsample frames during training and use every third frame to avoid redundancy. All models are trained end-to-end using the AdamW optimizer \cite{loshchilov2018decoupled_adamw} with a learning rate of 1e-4 for our experiments on GazeFollow, and a learning rate of 1e-5 for our experiments on VideoAttentionTarget. The loss coefficients are set to 100 for $\lambda_{gaze}$, 0.1 for $\lambda_{dir}$, and 1 for $\lambda_{io}$ and $\lambda_{att}$. We train for 35 epochs on GazeFollow, and 20 epochs (40 for the Multimodal model) on VideoAttentionTarget.

\mypartitle{Performance Metrics.}
The typical metrics used to evaluate  gaze target prediction are:
\begin{compactitem}
    \item \textbf{AUC}: The predicted gaze target heatmap is compared against a binarized version of the ground truth gaze target heatmap. This is used to plot a curve for the True Positive Rate vs the False Positive Rate. The AUC is the area under this curve, where 1 is perfect performance and 0.5 is random behavior.
    
    \item \textbf{Distance}: The predicted gaze location is compared against the ground truth  location using an L2 distance. We assume that each image is of size $1 \times 1$ when computing the L2 distances. Hence, distance values range from 0 to $\sqrt 2$, where a lower value is better.
    When multiple annotations are available  for the gaze location (GazeFollow), we compute the minimum and average distances to aggregate across all ground-truth labels.
    
    \item \textbf{Average Precision} (AP) is used to evaluate classification performance for the in vs out of frame prediction.
\end{compactitem}
%
The AP is computed across the entire test set, and the distance and AUC on the subset of images with a ground-truth gaze target located inside the frame.

\subsection{Tested models}
\label{sec:testedmodels}


\mypartitle{Individual modalities.}
To evaluate the strength of each modality, we tested the model by relying on a single modality as input. In this case, the scene module does not include the fusion mechanism, and the feature map 
$\featuremap$ of that modality is used directly as input to the Prediction module.

\mypartitle{Privacy approach.}
In this approach, the goal is to rely only on processed and anonymized input data. Concretely, the Human-Centric module takes as input the crop of the subject's head from the pose image rather than the input image. The head skeleton is treated as an RGB input, and no changes are made to the architecture or training protocol.

\mypartitle{Late fusion.}
We also evaluate a late fusion scheme where the gaze cone image $g_{cone}$ and the binary head mask $h_{loc}$ are fused with the Scene-Centric stream later in the architecture. Specifically, the two images are first downsampled before being concatenated together with the feature map $\featuremap_m$ of each modality separately. This is represented by the blue dashed line in Figure \ref{fig:full-model}.

\mypartitle{Skip connections.}
To evaluate the importance of retaining information from higher resolutions during the upsampling process in the decoder of the Scene-Centric branch, we train a model on the image alone, without skip connections.

\mypartitle{Modality Dropout.}
To evaluate the importance of modality dropout during training, we train a multimodal model without modality dropout.

\mypartitle{State-of-the-art.}
We compare the performance of our approach to different state-of-the-art methods for this task. Specifically, we include models from Chong et al. \cite{chong2020dvisualtargetattention},
Lian et al. \cite{lian2018believe}, Jin et al. \cite{jin2021multi} and Fang et al. \cite{Fang_2021_CVPR_DAM}.
Given that some works use a temporal variant of their model on VideoAttentionTarget, we include their static variant as well for the sake of fairness when comparing the results.

\subsection{Results}
\begin{table}[t]
    \centering
    \begin{tabular}{l c c c}
    \hline
    \textbf{Model} & \hspace{-2mm}\textbf{AUC}$\uparrow$ & \hspace{-2mm}\textbf{AvgDist}$\downarrow$ & \hspace{-2mm}\textbf{MinDist}$\downarrow$ \\
    \hline 
    \hline
    Lian~\cite{lian2018believe} & 0.906 & 0.145 & 0.081\\
    Chong~\cite{chong2020dvisualtargetattention} & 0.921 & 0.137 & 0.077\\
    Jin~\cite{jin2021multi} & 0.919 & 0.126 & 0.076 \\
    Fang~\cite{Fang_2021_CVPR_DAM} & 0.922 & \textcolor{blue}{0.124} & \textcolor{blue}{0.067}\\
    \hline
    \hline
    Image & \textcolor{blue}{0.933} & 0.134  & 0.071\\
    Depth & 0.921 & 0.141  & 0.080\\
    Pose & 0.902 & 0.164  & 0.100\\
    \textbf{Multimodal} & \textcolor{red}{0.943} & \textcolor{red}{0.114}  & \textcolor{red}{0.056}\\
    \hline
    \hline
    Depth-privacy & 0.920 & 0.152 & 0.088\\
    Pose-privacy & 0.893 & 0.175 & 0.109\\
    Multimodal-privacy & 0.928 & 0.136 & 0.075 \\
    \hline
    \hline
    Image-NoSkip & 0.932 & 0.133 & 0.073\\
    Multimodal-NoMoDrop & 0.941 & 0.115 & 0.057\\
    Multimodal-Late & 0.931 & 0.128  & 0.068\\
    \hline
    \end{tabular}
    \vspace*{-1mm}
    \caption{Results for our models on the GazeFollow dataset. Best scores are given in \textcolor{red}{red} and second best scores are given in \textcolor{blue}{blue}.}
    \label{tab:gazefollow-results}
\end{table}

\begin{table}[t]
    \centering
    \begin{tabular}{l c c c}
    \hline
    \textbf{Model} & \textbf{AUC} $\uparrow$ & \textbf{Dist} $\downarrow$ & \textbf{AP} $\uparrow$ \\
    \hline 
    \hline
    Chong~\cite{chong2020dvisualtargetattention}-static & 0.854 & 0.147 & 0.848\\
    Chong~\cite{chong2020dvisualtargetattention} & 0.860 & 0.134 & 0.853\\
    Jin~\cite{jin2021multi} & 0.870 & 0.127 & \textcolor{blue}{0.882}\\
    Fang~\cite{Fang_2021_CVPR_DAM} & 0.905 & \textcolor{red}{0.108} & \textcolor{red}{0.896}\\
    \hline
    \hline
    Image & \textcolor{red}{0.918} & 0.122 & 0.864\\
    Depth & 0.899 & 0.134 & 0.852\\
    Pose & 0.904 & 0.131 & 0.866\\
    \textbf{Multimodal} & \textcolor{blue}{0.913} & \textcolor{blue}{0.110} & 0.879\\
    \hline
    \hline
    Depth-privacy & 0.891 & 0.156 & 0.831\\
    Pose-privacy & 0.881 & 0.150 & 0.823\\
    Multimodal-privacy & 0.895 & 0.140 & 0.826\\
    \hline
     \hline
    Image-NoSkip & 0.906 & 0.133 & 0.857\\
    Multimodal-NoMoDrop & 0.905 & 0.118 & 0.874\\
    Multimodal-Late & 0.905 & 0.113 & 0.863\\
    \hline
     \hline
    \end{tabular}
       \vspace*{-1mm}
    \caption{Results on the VideoAttentionTarget dataset. Best scores are given in \textcolor{red}{red} and second best scores are given in \textcolor{blue}{blue}.}
    \label{tab:videoatt-results}
\end{table}


Our results on the GazeFollow and VideoAttentionTarget datasets are summarized in Table~\ref{tab:gazefollow-results} and  Table~\ref{tab:videoatt-results}.

\mypartitle{Individual Modalities.}
As the image contains the most complete information, 
it logically has the best performance on both datasets. 
Surprisingly, the performance of the depth and pose modalities are not very far, esp. when compared to state-of-the-art methods. 
In general, pose might be more accurate than depth when gaze is on faces or hands, but much worse when gaze is on scene objects since 
pose images contain absolutely no scene information.
We observe that according to all metrics, depth is better than pose on GazeFollow, while it is around the same on VideoAttentionTarget. 
This can be explained  by the fact that VideoAttentionTarget has more gaze points on faces compared to GazeFollow.

\mypartitle{Multiple Modalities.}
Our Multimodal model gives better results compared to using the image alone on both datasets.
The improvements are mainly visible on the distance metrics, where error reductions of 10\% to 21\% (the MinDist on GazeFollow) are achieved, which might be explained by the improved localization accuracy due to the pose cue, as well as disambiguation from the depth cue.   

Compared to the state-of-the-art, we can see that our approach performs  better than existing methods on GazeFollow for all metrics (e.g. reduction of more than 10\% on distance metrics compared to ~\cite{Fang_2021_CVPR_DAM}).
On VideoAttentionTarget, our results are in par with those of Fang \etal~\cite{Fang_2021_CVPR_DAM}.
As Fang et al.~\cite{Fang_2021_CVPR_DAM} process the eye regions to 
infer the gaze direction (which we do not), and eyes are  
more clearly visible in the VideoAttentionTarget dataset
(it contains higher resolution front facing faces), 
we hypothesize that adding such information in our model 
would further improve our results.

\begin{figure*}[tb]
    \centering
    \includegraphics[width=\textwidth]{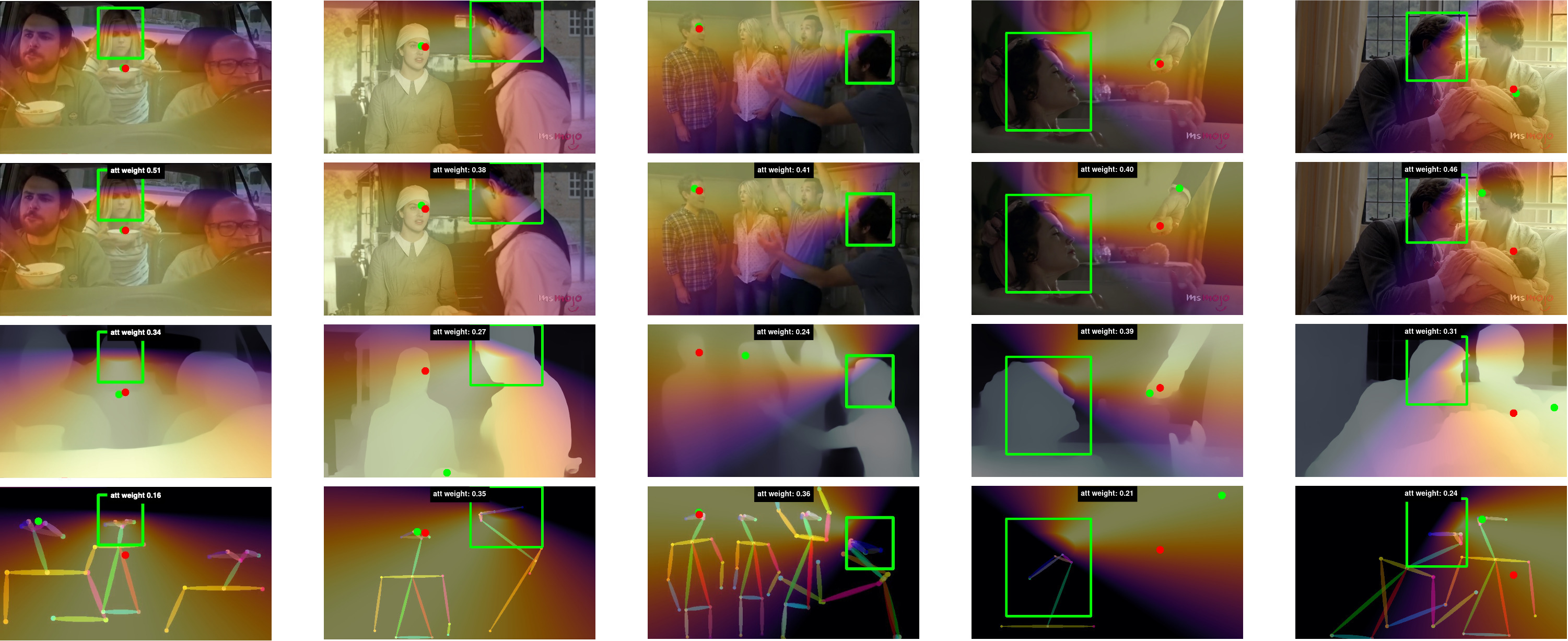}
    \vspace*{-5mm}
    \caption{Qualitative results of our  models (from top to bottom: multimodal, image, depth and pose). 
    The image (or modality) is superimposed with the predicted gaze cone, the predicted gaze target (in \textbf{\textcolor{green}{green}}) and the ground truth target (in \textbf{\textcolor{red}{red}}).
    We observe that the attention scores reflect the reliability of the respective modalities for a particular sample (pose in 2nd and 3rd column; depth in 4th column), and that the fusion is able to ignore wrong information (pose in 1st and 4th columns; depth in 3rd column), and improve predictions of the image modality (4th and 5th column). 
    }
    \vspace*{-3mm}
    \label{fig: qualitative}
\end{figure*}

\mypartitle{Qualitative examples and attention scores.}
Qualitative examples are provided in Figure~\ref{fig: qualitative}, 
where the outputs of the single and multimodal models are displayed. 
The attention scores predicted in the multimodal case are also indicated 
on each image modality for the given example.  
As can be seen, these scores reflect somehow the reliability associated with each cue. 
More generally, on GazeFollow, 
the average attention scores are 0.41, 0.36, 0.23 for the image, 
depth and pose modality, and 0.37, 0.31 and 0.32 on VideoAttentionTarget, 
reflecting the higher importance (and accuracy) of pose in this dataset as described earlier.

%

%

\mypartitle{Privacy setting.}
On GazeFollow, the depth and pose models have similar performance to their counterparts where the gaze direction is inferred from a head crop of the image  
rather than from the pose image. 
However, on VideoAttentionTarget, there is a drop of performance. 
We believe this is because in GazeFollow, eyes are less visible and 
there are many instances where the face is not visible (hence, the head pose 
is the only available cue). 
In contrast, VideoAttentionTarget contains higher resolution front-facing faces 
where eyes can bring additional gaze information.
The multimodal version (using depth and pose) improves performance (esp. on distance metrics) compared to the single modalities, obtaining similar performance to the image only model on GazeFollow and comparable performance to most state-of-the-art baselines on VideoAttentionTarget.

\mypartitle{Skip-connections.}
We believe that one reason for the superior performance of our models compared to the state-of-the-art is the use of skip connections during the upsampling process. Without this feature (NoSkip model in result tables), we observe that while the performance is unchanged on GazeFollow, there is a  performance drop on VideoAttentionTarget, esp. on the distance metrics.
This may be because the higher resolution images of VideoAttentionTarget 
contain higher details about the face (eyes, nose) or objects, and can thus 
benefit more from the skip connections to precisely regress the target gaze locations.  

\mypartitle{Late fusion.}
The late fusion of the gaze information (gaze cone) with the feature maps
rather than with the input images in our multimodal model improves performance 
compared to the models trained on a single modality (with early fusion), but has lower 
performance compared to the early fusion strategy in the multimodal case. 
We believe this is because introducing the person-specific gaze information early  
results in more capacity to identify the potential gaze target for that person. 
With the late fusion, the model has to identify potential  
gaze targets for all people in the scene (and at any place in the image). 

\mypartitle{Modality dropout.}
We analyze the importance of modality dropout during training. Without this feature, the multimodal model (NoMoDrop in the results tables) achieves a similar performance on GazeFollow, but obtains worse results on VideoAttentionTarget.

In the case of VideoAttentionTarget the average attention weights for image, depth and pose without modality dropout are 0.32, 0.29, 0.39, and the average weights with modality dropout are 0.37, 0.31, 0.32. Hence, modality dropout helps to learn a distribution of attention weights which better reflects the importance of the modalities. This may in turn help the model make better gaze target predictions.

\section{Conclusion}

In this paper, we proposed a modular multimodal architecture to explicitly leverage pose and depth information in order to improve the predicted gaze location and improve the state-of-the-art performance on two public benchmarks.
We also investigated a late fusion scheme which allows us to first parse the scene in a person-agnostic manner, before introducing the subject's information. We showed that our model can also benefit privacy-sensitive applications in which personally identifiable information cannot be exposed. In this case, our model operates on head skeletons together with the pose and depth maps, achieving competitive performance.

Our architecture is modular and can naturally be extended to include other modalities, 
like 
optical flow (for videos), which we believe can further improve predictions.
Alternatively, we can extend our model to inherently incorporate temporal information. 
Secondly, it is not clear at this point whether the depth cue is used as a way to verify the depth compatibility of the inferred gaze target with respect to the head position and gaze. Further study is needed to evaluate this. Finally, our current attention mechanism implies that one modality should be chosen to predict gaze. Conceptually, this formulation assumes the different modalities are equivalent, which is not necessarily the case. 
Thus, another future direction could investigate how to better fuse information across modalities.

\mypartitle{Acknowledgement.}
This research has been supported by the ROSALIS project (Robot skills acquisition through active learning and social interaction strategies, grant agreement no. 30214) of the Swiss National Science Foundation (SNSF) as well as by the AI4Autism project (Digital Phenotyping of Autism Spectrum Disorders in children, grant agreement no. CRSII5\_202235 / 1) of the the Sinergia interdisciplinary program of the SNSF.


{\small
\bibliographystyle{ieee_fullname}
\bibliography{MultimodalAttention}
}

\end{document}